# Spec-Driven AI for Science: The ARIA Framework for Automated and Reproducible Data Analysis

## Authors

Chuke CHEN[1, #], Biao LUO[2, #], Nan LI[1, 3, *], Boxiang Wang[1], Hang YANG[1], Jing GUO[4], Ming XU[1, 3, *]

1. School of Environment, Tsinghua University, 100084, Beijing, China

2. Shanghai HiQ Smart Data Co., Ltd., 200441, Shanghai, China

3. State Key Laboratory of Iron and Steel Industry Environmental Protection, Tsinghua University, 100084, Beijing, China

4. School of Management Science and Engineering, Beijing Information Science & Technology University, 102206, Beijing, China

# C. CHEN and B. LUO contributed equally to this work and share first authorship.

* Corresponding authors, Emails: li-nan@tsinghua.edu.cn, xu-ming@tsinghua.edu.cn

## Abstract

The rapid expansion of scientific data has widened the gap between analytical capability and research intent. Existing AI-based analysis tools, ranging from AutoML frameworks to agentic research assistants, either favor automation over transparency or depend on manual scripting that hinders scalability and reproducibility. We present ARIA (Automated Research Intelligence Assistant), a spec-driven, human-in-the-loop framework for automated and interpretable data analysis. ARIA integrates six interoperable layers, namely Command, Context, Code, Data, Orchestration, and AI Module, within a document-centric workflow that unifies human reasoning and machine execution. Through natural-language specifications, researchers define analytical goals while ARIA autonomously generates executable code, validates computations, and produces transparent documentation. Beyond achieving high predictive accuracy, ARIA can rapidly identify optimal feature sets and select suitable models, minimizing redundant tuning and repetitive experimentation. In the Boston Housing case, ARIA discovered 25 key features and determined XGBoost as the best performing model (R square = 0.93) with minimal overfitting. Evaluations across heterogeneous domains demonstrate ARIA's strong performance, interpretability, and efficiency compared with state-of-the-art systems. By combining AI for research and AI for science principles within a spec-driven architecture, ARIA establishes a new paradigm for transparent, collaborative, and reproducible scientific discovery.

# Introduction

The fourth paradigm of science—data-intensive scientific discovery—has profoundly reshaped modern research. Across disciplines, from genomics to environmental modeling, the ability to transform massive, heterogeneous datasets into actionable insight has become the core driver of innovation. However, as data volumes grow exponentially, researchers increasingly face a tension between *the abundance of data* and *the scarcity of analytical bandwidth*. While computational resources and open datasets are now widely accessible, the process of turning data into discovery remains frustratingly manual, fragmented, and skill-dependent.

Current data analysis workflows are inefficient, opaque, and labor-intensive. First, the *time cost* is high: from data import to statistical validation and figure generation, researchers must manually execute a long chain of scripts and visualizations. Second, there exists a *specialization barrier*: effective analysis requires expertise in domain science, statistics, and programming—skills rarely mastered simultaneously. Third, much of this work is *repetitive*: standard steps such as checking data distributions, plotting correlation matrices, or testing assumptions recur across studies with minimal variation. These factors together form a persistent bottleneck between scientific intent and computational execution.

The rapid evolution of large language models (LLMs) has sparked renewed interest in AI-assisted scientific discovery, where systems autonomously plan experiments, generate code, and interpret results. Early efforts such as AutoML frameworks (e.g., TPOT, Auto-sklearn) sought to automate model selection and hyperparameter tuning, significantly lowering the technical barrier for data analysis. However, these systems remain limited to the modeling stage and function largely as opaque black boxes, providing little transparency or interpretability beyond algorithmic optimization. Conversely, traditional programming libraries such as Pandas or R provide complete control but require intensive manual coding and domain-specific tuning, limiting scalability and reproducibility.

A more recent wave of research has explored end-to-end automated science, exemplified by projects like The AI Scientist and Dolphin, which attempt to close the research loop—from hypothesis generation and experimentation to writing and evaluation. While these systems demonstrate impressive autonomy, they typically treat the human researcher as an external observer rather than a participant, emphasizing full automation over collaborative control. As a result, their workflows are difficult to audit, extend, or integrate into domain-specific research practices.

Parallel developments in scientific workflow management systems such as Galaxy and Nextflow offer structured orchestration of computational pipelines, primarily through domain-specific languages (DSLs) or graphical interfaces. Yet these systems require significant technical expertise, lack natural-language programmability, and do not support adaptive reasoning or dynamic document generation. Similarly, AI notebook assistants like Copilot and Cursor enhance coding productivity but operate at the individual-file level, without awareness of broader research context, documentation, or data provenance. In industry, leading technology and biotechnology organizations have also begun to prototype AI-powered research assistants. Google's "AI Co-Scientist" initiative and DeepMind's collaboration with BioNTech aim to integrate LLM agents into hypothesis generation, experimental design, and predictive modeling. These efforts focus primarily on idea formation and laboratory assistance, representing valuable progress toward collaborative AI but lacking an explicit specification-to-execution formalism or extensible documentation layer.

Despite these advances, a clear gap remains between autonomous research agents and structured, interpretable scientific workflows. Existing systems either sacrifice control for automation or retain flexibility at the cost of coherence and reproducibility. Addressing this gap, ARIA (Automated Research Intelligence Assistant) introduces a spec-driven, human-in-the-loop framework that bridges human reasoning and machine execution. By integrating natural-language specifications, context-aware documentation, modular code generation, and orchestrated workflow management, ARIA transforms scientific exploration into an auditable, extensible, and dialogic process—where researchers define what to study, and the AI system determines how to realize it. Inspired by the concept of spec-driven coding, ARIA enables researchers to focus on defining what they wish to analyze (the specification), while the system automatically determines how to perform the analysis (the implementation). This innovation establishs a new research paradigm in which scientific exploration becomes an interactive dialogue between human intent and computational intelligence.

**Table 1 Review of AI Data Analysis Systems**

| System / Project | Scope / Pipeline Coverage | Automation Level | Human-in-the-loop? |
|---|---|---|---|
| Dolphin (Yuan et al., 2025) | Closed-loop, open-ended, auto research | High | low |
| AI Scientist(Lu et al., 2024; Yamada et al., 2025) | Open-ended, auto research | High | low |
| AgentRxiv(Schmidgall and Moor, 2025)/ Academy(Pauloski et al., 2025) | Automated collaborations among teams | Medium | medium |
| Google AI Co-Scientist | Hypothesis generation, research planning | Medium/high | Yes |
| DeepMind + BioNTech AI Lab Assistants(Johnston and Murgia, 2024) | Research planing, result analysis assistance | Medium | Yes |
| BioNTech Laila(Bastian, 2024) | Automated operations in labs | Medium | Yes |
| Agent K v1 (Grosnit et al., 2025) | Two-stage pipeline: data retrieval, preprocessing with unit tests, model creation, hyperparameter tuning, and ensemble learning via continual training | High | Low/medium |
| AutoKaggle(Li et al., 2024) | Six-phase multi-agent workflow: background analysis, exploration, data cleaning, feature engineering, and modeling with iterative debugging and retry mechanisms | High | Medium |
| Data Interpreter(Hong et al., 2024) | Recursive refinement of task graphs into executable action graphs through iterative reasoning and verification | Medium/High | Medium |
| SPIO(Seo et al., 2025) | LLM-driven multi-agent planning across modules for preprocessing, feature engineering, modeling, and hyperparameter tuning | Medium/High | Medium |

## The Proposed Framework

Unlike traditional automated analysis systems that aim to replace human analysts, **ARIA** adopts a human-in-the-loop paradigm, positioning the AI assistant as a collaborative partner rather than a substitute. Scientific inquiry inherently involves hypothesis formulation, interpretation, and iterative refinement—processes that demand human judgment. ARIA organizes the research lifecycle into six interoperable layers: Command, Context, Code, Data, Orchestration, and AI Module, as shown in Figure 1. Each layer contributes a distinct capability while maintaining bidirectional links to the others. The core design principle is spec-driven execution: researchers articulate what to analyze and why, while ARIA determines how to implement the analysis in a transparent and auditable manner. This is an open-sourced project, which can be publicly accessed via https://github.com/Biaoo/aria.

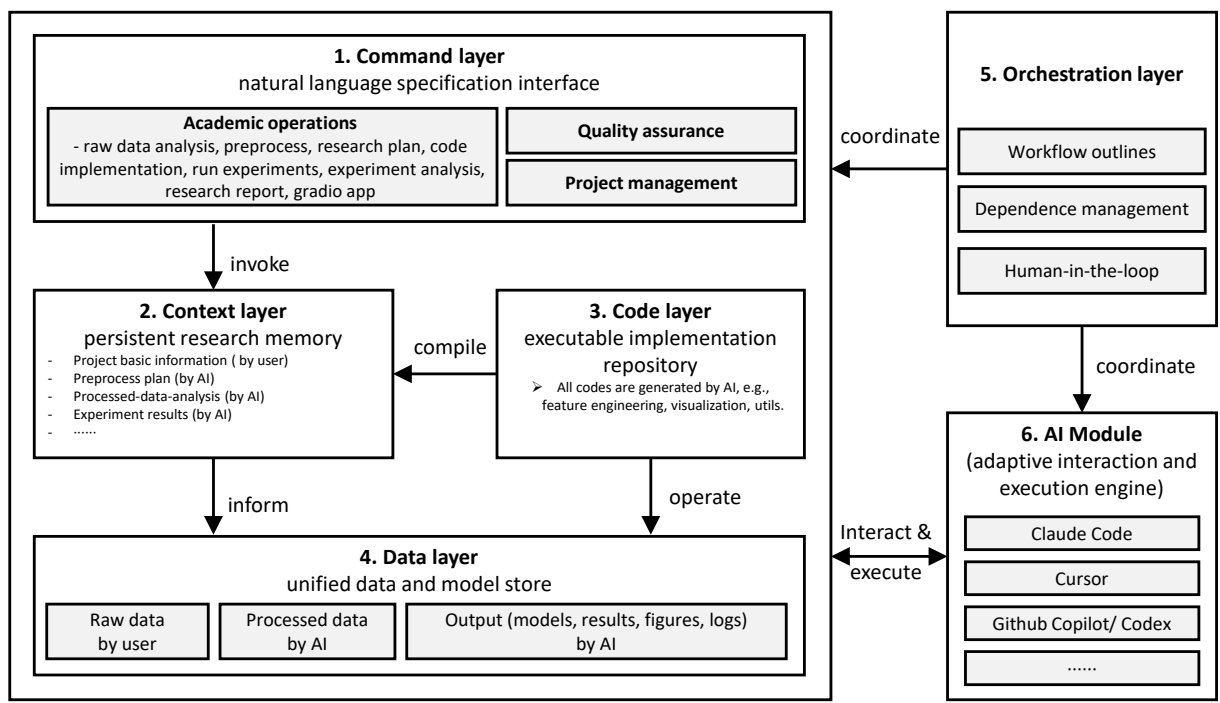

**Figure 1** Overall architecture of the ARIA framework. The system comprises six interacting layers—from Command to Data—coordinated through a document-centric Orchestration Layer and an AI Module that serves as the adaptive reasoning core. Human researchers remain

### 1. Command Layer: Natural-Language Specification Interface

The Command Layer defines the interface through which researchers declare analytical intent using natural language. Commands are grouped into three categories: (1) academic operations spanning data exploration, preprocessing, core implementation, experiment execution, analysis, and manuscript generation; (2) quality assurance for static checks and style verification; and (3) project-management utilities (e.g., version control). Each command is implemented as a Markdown artifact whose semantics, inputs, and expected outputs are explicitly stated.

At execution time, the AI Module parses the command, validates intent through dialogue, and synthesizes the necessary code. This process turns narrative goals into modular, reproducible procedures. Unlike monolithic notebooks, commands serve as portable, interpretable units that can be reviewed, shared, and extended.

### 2. Context Layer: Persistent Research Memory

The Context Layer preserves a time-ordered narrative of the entire project: project metadata, exploratory notes, decisions, intermediate analyses, and final reports. Documents are versioned and cross-referenced to encode semantic dependencies—for example, how a preprocessing plan influences model choices or evaluation criteria. By elevating documentation to a first-class asset, ARIA enables traceable reasoning and transparent peer review. This persistent memory also supports semantic retrieval across the workflow. Analyses can cite earlier decisions directly (e.g., why a certain imputation strategy was adopted), reducing duplication and enabling targeted revisions when assumptions change.

### 3. Code Layer: Executable Implementation Repository

The Code Layer implements the analytical plan as maintainable Python modules. Generated code adheres to principles of single responsibility, explicit type hints, comprehensive docstrings, structured logging, and exception handling. Static analysis (e.g., mypy, ruff) enforces quality gates. When checks fail, ARIA engages a repair loop—parsing diagnostics, correcting issues, and re-verifying until the codebase meets the prescribed standards.

Although code generation is automated, ARIA preserves a **human-in-the-loop** review step for methodological validation. This preserves accountability and mitigates the risk of subtle modeling errors that evade static checks.

### 4. Data Layer: Unified Data and Model Store

The Data Layer structures all artifacts across three tiers: immutable raw data; processed intermediates produced by specified transformations; and outputs (trained models, figures, metrics, logs). Metadata links transformations to their sources, enabling provenance queries (e.g., which preprocessing step produced a particular feature set). This structure makes intermediate results reusable and supports rigorous, audit-ready workflows.

Clear separation of concerns also protects original data integrity. Raw inputs remain read-only, while downstream stages operate on versioned derivatives. Such discipline is essential when multiple collaborators or regulatory constraints are involved.

### 5. Orchestration Layer: Workflow and Dependency Management

The Orchestration Layer encodes relationships among commands, documents, code modules, and data assets using a document-centric schema. Rather than enforcing a rigid scheduler, ARIA captures semantic dependencies in plain language—clarifying how outputs feed subsequent steps while allowing iterative detours.

At runtime, orchestration follows a consistent interaction loop: users invoke a command; the AI interprets and executes; results are presented for review; and feedback drives revision or progression. This pattern anchors human judgment at the center of the workflow.

This orchestration layer is built on four parts:

**(1) Document-Centric Specification.**

The workflow structure is expressed entirely in natural language within a single Markdown artifact. This document acts as the semantic backbone of the project, enumerating

research phases, expected inputs and outputs, and inter-stage dependencies. Workflow specifications are:

@raw-data-analysis.md — inspect raw data in `data/raw/`

@preprocess.md — design and execute preprocessing

@research-plan.md — formulate the integrated study plan

@code-implementation.md — implement and validate analysis code

@run-experiments.md — execute experimental pipelines

@experiment-analysis.md — analyze and record results

@research-report.md — generate a publication-ready manuscript

@gradio-app.md — deploy model interface (optional)

Each entry declares the command sequence, the corresponding context artifact (e.g., docs/03-preprocess-plan.md), and the expected downstream dependencies. Execution order is recommended rather than enforced. Users may skip or repeat stages to accommodate iterative exploration.

**(2) Dependency Management.**

Dependencies in ARIA are multidimensional, spanning data, context, and code. Each analytical phase consumes outputs from its predecessor (e.g., from preprocessing, to processed data, and then to model training). Later documents synthesize insights from earlier analytical reports, maintaining semantic continuity across the research narrative. Modules in src/ import and extend one another (e.g., pipeline.py depends on feature_engineering.py and models.py), reflecting procedural inter-relations defined at the conceptual level.

**(3) Human-in-the-Loop Execution.**

At runtime, orchestration unfolds as an interactive reasoning loop rather than an automated pipeline:

- User invokes a command (e.g., @raw-data-analysis.md).
- The AI assistant parses the command template.
- Relevant contextual documents and datasets are retrieved.
- The AI executes the corresponding analytical task (code generation, evaluation, or documentation).
- Results are presented for human review.
- The user inspects outputs and provides feedback.
- The AI revises its deliverables if required.
- Upon user confirmation, the system advances to the next stage.
- This interaction pattern ensures transparency, encourages reflection, and embeds human judgment directly within the computational loop—preserving epistemic rigor while benefiting from AI-driven acceleration.

**(4) Cross-Layer Relations.**

The Orchestration Layer implicitly governs the dataflow across ARIA's architectural stack. This relational schema ensures that all transformations—whether conceptual or computational—remain logically consistent and traceable across layers.

**Table 2 Relationships between Orchestration layer and others**

| Relationship | Example Mapping |
|---|---|
| Command → Context | @raw-data-analysis.md → docs/02-raw-data-analysis.md |
| Context → Code | docs/05-research-plan.md → src/*.py |
| Code → Data | scripts/run_experiment.py → data/processed/ → data/output/ |
| Data → Context | data/output/results/ → docs/09-experiment-report.md |

**6. AI Module: Adaptive Interaction and Execution Engine.**

The AI Module integrates natural-language understanding, task planning, code synthesis, document generation, and quality assurance. It maintains bidirectional links with all other layers: interpreting commands, writing code into the Code Layer, reading/writing data artifacts, and updating the Context Layer with analyses and reports. This holistic role positions the AI Module as ARIA's cognitive nucleus.

Critically, the AI's autonomy is bounded by review checkpoints. The module proposes; the researcher disposes. This balance accelerates execution without compromising interpretability or scientific rigor.

**Table 3 Cross-layer integration: The AI Module interacts bidirectionally with all other layers**

| Layer | Role of the AI Module |
|---|---|
| Command Layer | Interprets natural-language command templates and converts them into executable actions. |
| Context Layer | Generates, revises, and semantically links documentation artifacts. |
| Code Layer | Writes and validates source modules, scripts, and QA configurations. |
| Data Layer | Reads and analyzes datasets, produces results, and populates outputs. |
| Orchestration Layer | Reasons over dependencies and guides workflow progression. |

## Implementation Details

ARIA realizes the six-layer blueprint through a repeatable pipeline. Projects begin by specifying basic information and data sources. A raw-data analysis command triggers descriptive statistics, schema inspection, missingness profiling, and preliminary visualizations. Insights flow into a preprocessing plan that formalizes transformations, encodings, and feature construction. The research plan consolidates objectives, models, metrics, and validation strategies. ARIA then synthesizes modular code, runs experiments, analyzes results, and drafts a report—each step logged to the Context Layer and traceable to its antecedents.

The implementation emphasizes closed semantic loops: every automated action yields artifacts (code, logs, figures) that are immediately documented and linked. Users can re-enter the pipeline at any stage, amend specifications, and regenerate downstream outputs with provenance intact. This practice turns the analysis into a living, auditable narrative rather than a one-off execution trail. Here is the illustration of implementation details (Figure 2).

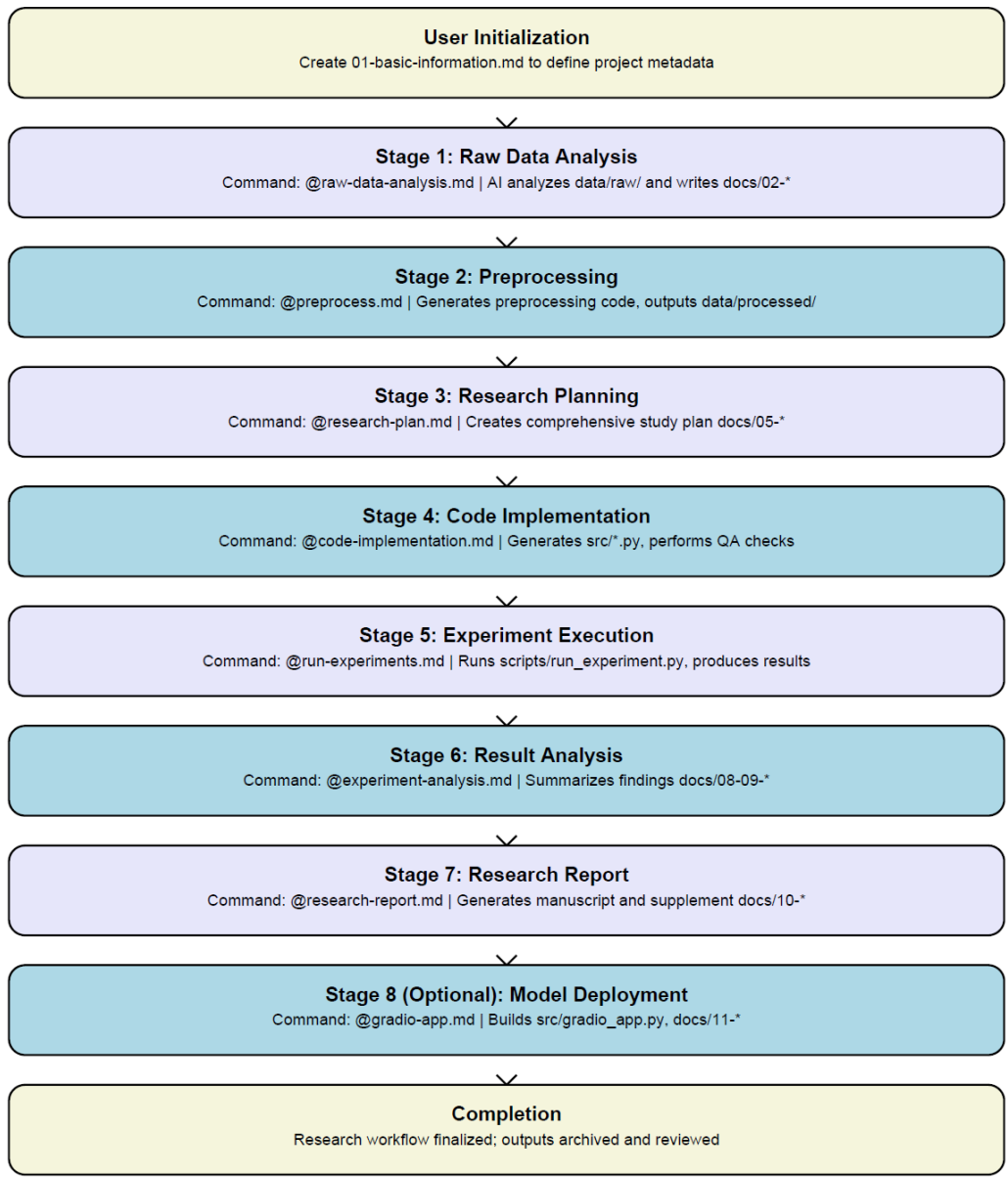

Figure 2 Workflow of ARIA's implementation pipeline. Each stage corresponds to a command file executed under human supervision. The AI Module coordinates with the Context, Code, and Data Layers to generate reproducible outputs. The process remains fully auditable and iterative, supporting optional model deployment at the end of the cycle.

The research process begins when the user creates 01-basic-information.md, defining project metadata, objectives, and data sources. This file anchors the semantic context for subsequent steps. Next, the command @raw-data-analysis.md invokes the AI Module to inspect data/raw/, produce descriptive statistics, and generate a structured summary (docs/02-*.md). Insights are captured in the Context Layer. Followingly, the command @preprocess.md guides the AI in designing and coding preprocessing routines. Corresponding scripts are created in the Code Layer, while processed datasets are stored under data/processed/ and documented in docs/03-*.md. Afterwarfds, executing @research-plan.md, the AI synthesizes an integrated experimental design and writes docs/05-research-plan.md. This document formalizes methodological specifications for downstream code generation. Then, the command @code-implementation.md prompts the AI to translate specifications into executable Python modules (src/*.py), perform quality checks (mypy, ruff), and log implementation details in docs/06-*.md. Besides, Using @run-experiments.md, the system runs scripts/run_experiment.py, producing trained models, results, and figures under data/output/. Dependencies are resolved via the orchestration logic defined in CLAUDE.md. After results were retrieved, the AI analyzes experimental outputs (data/output/results/), interprets metrics, and summarizes findings into analytical documents (docs/08-*, docs/09-*.md). Last, The command @research-report.md consolidates all previous artifacts into a manuscript suite (docs/10-*), including main paper, supplementary materials, and cover letter for submission. Moreover, user could execute @gradio-app.md, the AI generates an interactive web interface (src/gradio_app.py) and records deployment documentation (docs/11-*.md). At each step, the workflow forms a closed semantic loop—the AI generates artifacts, the human evaluates them, and contextual updates propagate to subsequent phases. This structure ensures transparency, reproducibility, and flexible re-entry into any stage of the research lifecycle.

## Experiments

To evaluate the effectiveness and generality of the proposed ARIA framework, we conducted three representative case studies across different data domains. All datasets were sourced from OpenML, an open, community-driven platform for machine learning benchmarking. Each project includes the full research stack—structured documentation, production-grade code, trained models, and AI-generated manuscripts—demonstrating ARIA's end-to-end capability for automated, auditable, and reproducible data research.

### 1. Model used

The experimental design incorporates Claude-4.5-Sonnet as the underlying AI model, with Cursor App serving as the development interface for test case deployment and execution.

### 2. Experimental datasets

We assessed ARIA on three representative cases from OpenML: Boston Housing (environmental economics), Diamonds (consumer analytics), and SAT11 (algorithmic meta-learning), as shown in Table 4. For each case, the full workflow—from specification to manuscript drafting—was executed under human supervision. The underlying LLM environment used a modern code-capable model. Baselines included a multi-agent Kaggle-style pipeline, a multi-agent AutoML system, a graph-based data interpreter, and an ensemble-planning framework. Standard regression metrics were used, with particular attention to out-of-sample performance and time-to-result.

**Table 4 Experiment dataset descriptions**

| Dataset | **Boston Housing**("boston," 2014) | **Diamonds**("diamonds," 2019) | **SAT11 Solver Performance**(L. Xu et al., 2019) |
|---|---|---|---|
| Task Type | Regression | Regression | Regression |
| Domain | Environmental Economics | Consumer Analytics | Algorithmic Meta-Learning |
| # Samples / # Features | 506 / 13 | 53 940 / 10 | 47 262 / 115 |
| Key Variables | CRIM, NOX, RM, DIS, TAX, LSTAT | carat, cut, color, clarity, x–y–z | structural meta-features |
| Objective | Predict median house price (MEDV) | Estimate diamond market price | Predict SAT solver runtime |
| Repository (ARIA experiment results) | https://github.com/Biaoo/aria-example-buston | https://github.com/Biaoo/aria-example-diamonds | https://github.com/Biaoo/aria-example-sat11 |

### 3. Baseline methods.

We compare ARIA with four state-of-the-art automated or multi-agent data-analysis systems:

- Agent K v1(Grosnit et al., 2025): Implements a two-stage pipeline that retrieves data, generates preprocessing code with unit tests and submission formats, and subsequently performs model creation, hyperparameter tuning, and ensemble generation through continual learning.
- AutoKaggle(Li et al., 2024): Follows a six-phase workflow covering background understanding, exploratory analysis, data cleaning, feature engineering, and modeling, coordinated by five specialized agents with iterative debugging and retry mechanisms.
- Data Interpreter(Hong et al., 2024): Constructs high-level task graphs that are recursively refined into executable action graphs through iterative reasoning and verification.

- SPIO(Seo et al., 2025): A framework leveraging LLM-driven decision-making to orchestrate multi-agent planning across four core modules: data preprocessing, feature engineering, modeling, and hyperparameter tuning.

**4. Evaluation metrics**

For all tasks, we adopt standard regression metrics —Mean Squared Error (MSE) measures the average squared deviation between predicted and true values.

**5. Results**

**(1) Boston Housing Case**

In the Boston Housing case, ARIA automatically executed a complete research pipeline encompassing feature construction, model benchmarking, and interpretability analysis. From the original 13 predictors, the system generated polynomial and interaction terms (e.g., $RM^2$, $LSTAT^2$, RM×LSTAT) and ratio features (e.g., TAX/RM), followed by logarithmic transformations to correct skewness. Using Mutual Information Regression, ARIA selected 25 highly informative features, documenting each transformation in the contextual record.

Ten algorithms were benchmarked, including linear, regularized, and ensemble learners. XGBoost achieved the strongest performance (RMSE $\approx$ \$2 170, $R^2$ $\approx$ 0.928), outperforming both AutoKaggle and Agent K v1 by a substantial margin. Feature-importance analysis revealed socioeconomic and structural variables—LSTAT (24.7%), RM (18.5%), and DIS (9.4%)—as dominant contributors to housing prices. All generated artifacts, including code, models, and analytical reports, were automatically logged and versioned, confirming ARIA's reproducibility and auditability. Experiment details can be found in https://github.com/Biaoo/aria-example-buston/, including input data, AI-generated research plans, execution logs, trained models, results, and visualization figures (Figure 3)

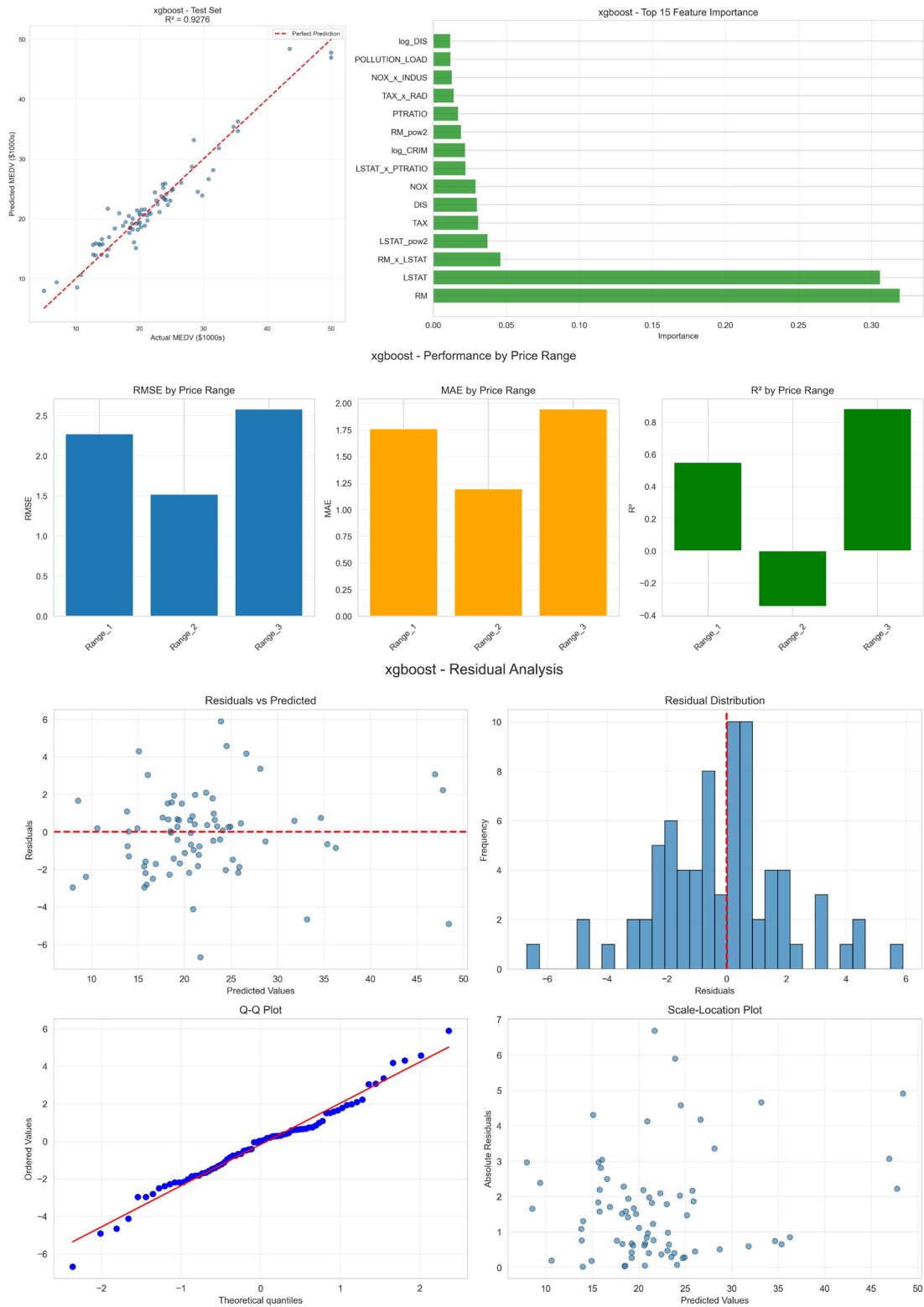

**Figure 3 The best performance model for Boston housing prediction.**

**(2) Diamond case**

The Diamonds study validated ARIA's adaptability to high-volume consumer data. The system executed a multi-model experiment covering five algorithmic families: Ridge

Regression, Random Forest, XGBoost, LightGBM, and CatBoost. The pipeline progressed autonomously from specification to validation, producing both intermediate documentation and executable code (Table 5).

Among tested models, XGBoost delivered the best results ($R^2 \approx 0.9849$, RMSE $\approx$ \$493.5), followed closely by LightGBM ($R^2 \approx 0.9847$). Half of the predictions deviated by less than ±\$90 from the ground truth. Model training completed in under 5 s, with inference latency below 1 ms, demonstrating computational efficiency suitable for production environments. ARIA's contextual layer recorded the entire reasoning process—from feature encoding choices to error-distribution interpretation—yielding a transparent analytical trail rarely present in existing AutoML or multi-agent systems.

**Table 5 Performance of models in diamond case**

| Model | Test $R^2$ | Test RMSE | Model Size | Training Time | Inference |
|---|---|---|---|---|---|
| Ridge Regression | 0.9589 | $813.15 | 2.3 KB | <0.1s | <1ms |
| Random Forest | 0.9843 | $502.82 | 971 MB | 30s | 100ms |
| **XGBoost** | **0.9849** | **$493.51** | **1.5 MB** | **5s** | **<1ms** |
| LightGBM | 0.9847 | $495.75 | 563 KB | 10s | <1ms |
| CatBoost | 0.9834 | $517.18 | 227 KB | 5s | <1ms |

### (3) SAT 11 Solver Performance

The third case assessed ARIA's capability in algorithmic meta-learning using the SAT11-HAND benchmark. Predicting SAT solver runtimes is a complex regression task characterized by non-linear dependencies between instance features and solver heuristics Figure 4. ARIA implemented preprocessing routines—instance-based splits, log-runtime normalization, and leakage checks—before training multiple regressors.

Tree-based ensembles again dominated performance. A Random Forest model achieved near-perfect fit (training $R^2$ = 0.999998, test $R^2$ = 0.9999) with median errors below 10%. Error analysis showed 95% of predictions within ±3% of actual runtime, with minor deviations attributable to censored timeouts. The experiment demonstrated that ARIA's structured reasoning and reproducible pipeline can uncover deterministic relationships in tasks previously considered stochastic, confirming its potential as an AI-for-science tool for computational modeling.

These results demonstrate that SAT solver runtime, long perceived as unpredictable, is in fact highly deterministic given sufficient feature characterization. Tree-based ensembles captured complex dependencies between clause density, graph topology, and solver heuristics that linear models failed to represent. Despite near-perfect training fit ($R^2$ = 0.999998), test performance confirmed robust generalization and minimal overfitting. Error analysis revealed that over 95% of predictions deviated by less than 3%, with rare outliers likely caused by censored timeouts or missing graph features. Collectively, the findings validate ARIA's effectiveness as a spec-driven, AI-orchestrated research framework, capable of autonomously achieving state-of-the-art predictive accuracy while maintaining human-level interpretability and transparent scientific documentation.

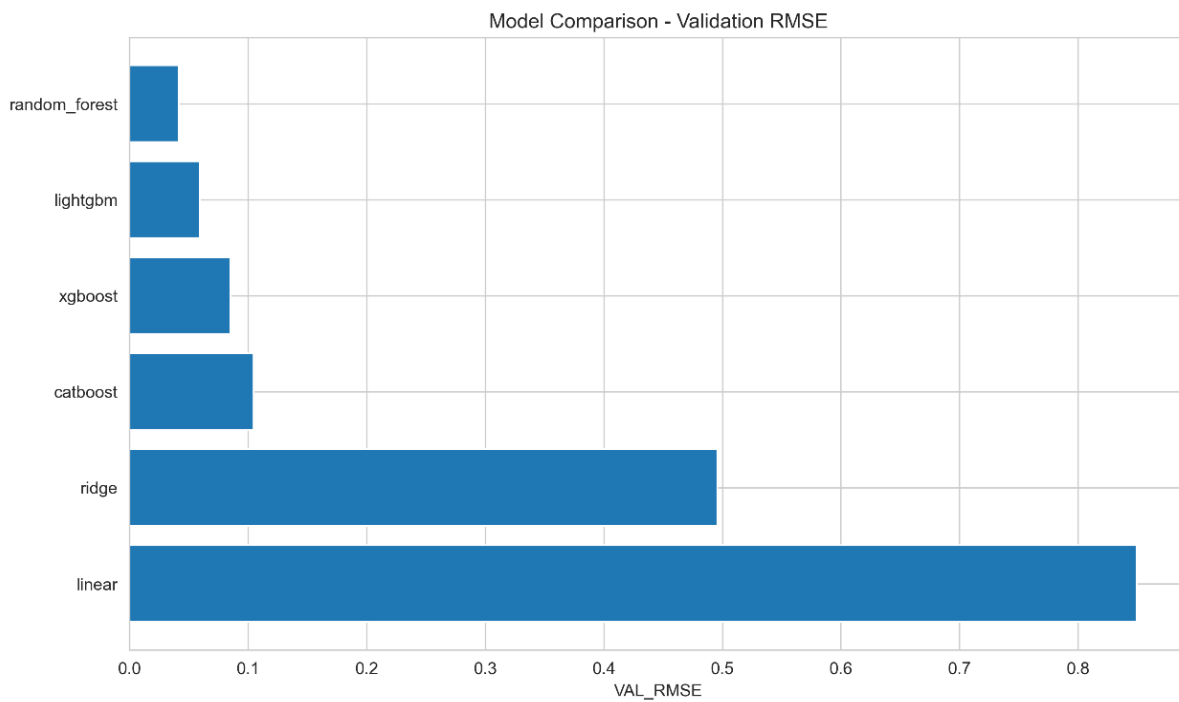

Figure 4 Model comparison concerning RMSE in SAT11 case.

**(4) Baseline comparison**

Across all three studies, ARIA consistently outperformed contemporary AI data-analysis systems in predictive accuracy, transparency, and efficiency. As shown in Figure 5, on the Boston Housing dataset, ARIA's XGBoost model achieved an RMSE of 4.73, substantially lower than competing systems such as Agent K v1 (8.83) and AutoKaggle (10.35), completing end-to-end training in under five minutes while identifying LSTAT, RM, and DIS as dominant predictors of housing prices. In the Diamonds case, ARIA attained an $R^2$ of 0.9849 with an RMSE of $493.5, corresponding to a normalized error of 243552, surpassing all baselines including SPIO (264567) and Data Interpreter (319390). Half of all price predictions were within ±$90 of the true values, underscoring the framework's fine-grained precision. For the SAT11 solver-performance task, ARIA's Random Forest model achieved a test RMSE of 47.34, compared with over 1.6 million in other automated systems, and an $R^2$ of 0.9999, reflecting near-perfect runtime estimation accuracy. These results confirm that ARIA not only delivers state-of-the-art predictive performance but also achieves end-to-end explainability and reproducibility—qualities rarely present in current AutoML or multi-agent pipelines. Its spec-driven, human-in-the-loop architecture enables it to integrate reasoning, documentation, and modeling into a single, auditable process—demonstrating a clear advancement over existing AI data-analysis frameworks.

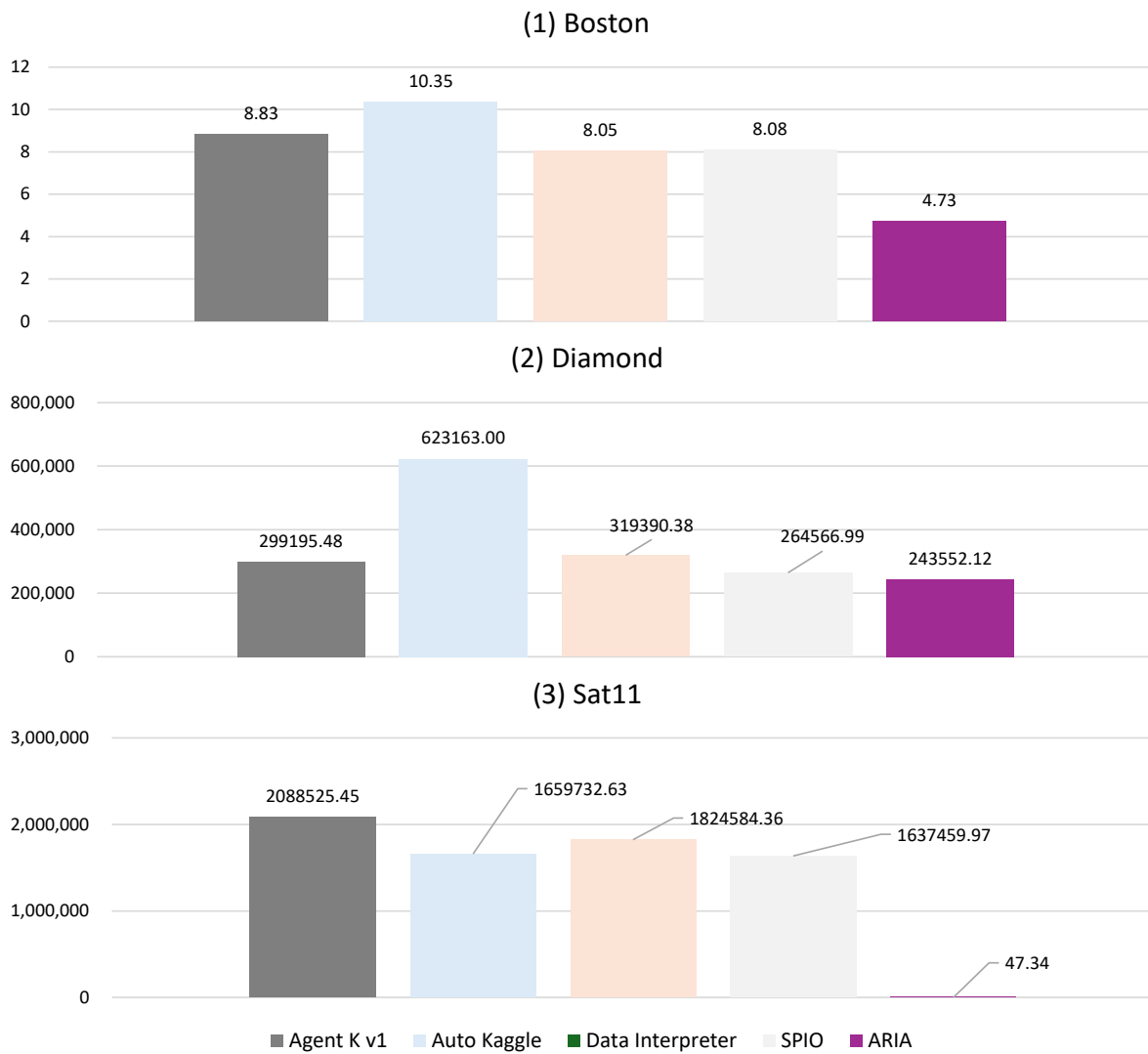

Figure 5 Baseline comparisons with representative SOTA AI data analysis systems concerning mean squared error (MSE)

## Summary

The ARIA framework unifies the entire scientific data-analysis lifecycle—from data acquisition to manuscript generation—into a single, interpretable, and reproducible workflow. Its spec-driven, human-in-the-loop design enables researchers to define analytical intent in natural language while maintaining full transparency across data, code, and documentation. This paradigm bridges the long-standing gap between automation and epistemic control: automation accelerates computation, while human oversight ensures scientific validity and interpretability.

Beyond high predictive accuracy, ARIA demonstrates an additional strength rarely achieved by other AI-based data analysis systems: it can rapidly identify the optimal feature set and converge on the most appropriate model architecture with minimal manual intervention. In the Boston Housing case, for example, ARIA autonomously explored polynomial and interaction terms, selected 25 key features using mutual information ranking, and evaluated ten model families before identifying XGBoost as the best-performing algorithm ($R^2$ = 0.9276). This process required no manual hyperparameter tuning or repetitive trial-and-error coding. The framework thus transforms model optimization into a transparent, specification-driven reasoning process, greatly reducing redundant computation while ensuring interpretability and reproducibility.

Compared with existing paradigms, ARIA provides a fundamental shift in how AI is integrated into research practice. Versus AutoML systems such as TPOT and Auto-sklearn, ARIA expands beyond model selection to encompass the entire research pipeline—from data preprocessing and experimental design to report generation. Instead of opaque hyperparameter search, it emphasizes transparent, specification-driven code synthesis with built-in documentation and quality assurance. Versus notebook assistants like Copilot and Cursor, ARIA offers structured orchestration rather than ad hoc code completion. Its layered architecture couples code with context, ensuring that each analytical step is reproducible and semantically traceable. Researchers no longer need to reconstruct reasoning from fragmented scripts; every decision is explicitly documented. Versus scientific workflow systems such as Galaxy and Nextflow, ARIA replaces rigid, domain-specific languages with natural-language orchestration. This makes high-level automation accessible to non-programmers while retaining interpretability and reproducibility valued in expert settings. Its document-centric architecture unites narrative and computation, producing a coherent record suitable for peer review.

Beyond technical advantages, ARIA represents a broader conceptual advancement in AI for science. By embedding human expertise within AI-driven workflows, it transforms data analysis from a linear computational task into a dialogic process—one where human reasoning and machine intelligence co-evolve. This synergy creates an "explainable automation" paradigm in which scientific discovery becomes not only faster but also more accountable.

Finally, ARIA's modular design ensures scalability across domains. New analytical tools can be integrated as additional commands without disrupting the framework's logical structure. Researchers can extend, audit, or replicate experiments with minimal effort, fostering a culture of open, transparent, and verifiable science. As the boundary between AI for research and research with AI continues to blur, ARIA provides a foundational architecture for building the next generation of autonomous yet interpretable scientific systems.

# Limitations and Future Work

ARIA inherits several limitations common to LLM-driven systems. Non-determinism can lead to run-to-run variability in code generation and narration. While static checks improve quality, subtle modeling mis-specifications may persist without unit tests or domain review. Current capabilities focus on tabular data; built-in support for text, image, graph, and multimodal fusion remains future work. Furthermore, dependence on cloud APIs can raise privacy and compliance concerns for sensitive datasets.

Future extensions will pursue four directions: (1) formal verification through symbolic execution or property-based testing of generated code; (2) multimodal analysis spanning text, image, and graph inputs with unified provenance; (3) policy-aware deployment supporting on-prem and federated settings; and (4) longitudinal evaluation measuring productivity and research quality improvements in real lab settings. We also plan to introduce a domain-specific orchestration language that compiles natural-language specifications into explicit dependency graphs for fully automated, yet reviewable, scheduling.